\documentclass[sigconf, nonacm=false]{acmart}

\setcopyright{acmlicensed}
\copyrightyear{2026}
\acmYear{2026}
\acmConference[RecSys '26]{20th ACM Conference on Recommender Systems Industry Track}
  {September 28--October 2, 2026}{Minneapolis, MN, USA}
\acmBooktitle{20th ACM Conference on Recommender Systems Industry Track (RecSys '26), September 28--October 2, 2026, Minneapolis, MN, USA}
\acmDOI{10.1145/nnnnnnn.nnnnnnn}
\acmISBN{978-x-xxxx-xxxx-x/YY/MM}

\usepackage{booktabs}
\usepackage{microtype}
\usepackage{tikz}
\usetikzlibrary{shapes.geometric, arrows.meta, positioning, fit, backgrounds, calc}

\makeatletter
\providecommand{\@LN}[2]{}
\providecommand{\@LN@col}[1]{}
\makeatother

\usepackage{seqsplit}  
\begin{document}

\title[Pairwise Ranking Outperforms Single-Action RL]
      {Pairwise Ranking Outperforms Single-Action RL for Offline Explanation Selection: A Practical Lesson}

\author{Tanay Chowdhury}
\authornote{Both authors contributed equally to this research.}
\affiliation{%
  \institution{Amazon}
  \city{Seattle}
  \state{WA}
  \country{USA}}
\email{tanaycho@amazon.com}

\author{Saeideh Shahrokh Esfahani}
\authornotemark[1]
\affiliation{%
  \institution{Amazon}
  \city{Mountain View}
  \state{CA}
  \country{USA}}
\email{saeidesh@amazon.com}

\renewcommand{\shortauthors}{Chowdhury and Esfahani}

\begin{abstract}
Industrial explainable-recommendation systems built around large
language models incur a substantial serving cost: each request
triggers an LLM generation, with latency in the hundreds of
milliseconds and per-query cost that scales linearly with traffic. We separate generation from
selection. Explanations are produced ahead of time as a frozen
candidate pool, six prompt styles applied to two commodity LLMs,
and at request time a small CPU-resident selector picks one. The
serving stack runs without GPUs and returns in under 100\,ms.

Our primary benchmark is a 2{,}958-pair XRec Google Local subset,
where we evaluate six offline-pool selectors (LambdaRank, PPO,
GRPO, DPO, and two-stage teacher--student distillation) and three
KG-path selectors (temperature-biased random walks, edge-disjoint
enumeration, and MMR-reranked paths with dual-style generation).
A 300-pair MovieLens-1M split with Claude-Sonnet-4.5-generated
references is used as an internal cross-dataset consistency
check; because no public explainable-recommendation benchmark
exists for MovieLens-1M, those numbers should be read as
intra-paper consistency evidence rather than as a second
benchmark. Every variant is scored against the same BERTScore-F1
protocol used by XRec and G-Refer, with results averaged across
five seeds.

LambdaRank reaches F1 = 0.500 on Google Local, exceeding both
G-Refer and XRec on the same 2{,}958-pair subset, and reaches
F1 = 0.329 on the MovieLens-1M consistency check. With seed
variance below 0.003 F1 per RL method, the ordering is
statistically reliable. The finding is consistent across both
data points: in this dense-label one-step bandit setting,
pairwise learning-to-rank outperforms the single-action RL
formulations (PPO, GRPO, DPO), which sample one labelled
candidate per rollout and leave the other $K{-}1$ labels out of
the gradient.

The KG-path family targets a different objective. All three of its
variants reach USR = 1.000 on Google Local and 0.997--1.000 on
MovieLens-1M, since per-request path grounding produces a unique
output per query and avoids the template-collapse failure mode
that can affect cached-LLM outputs.

A generator-pool study comparing Claude 3 Haiku and Claude
Haiku 4.5 shows small but measurable F1 shifts (0.001--0.006
across methods, with the larger end of that range exceeding the
across-seed standard deviation), while the relative ranking of
selectors is preserved. The qualitative design implication is
that selector and generator can be evaluated independently;
absolute F1 does depend on the generator. End-to-end build cost
is close to \$15 on commodity hardware.
\end{abstract}



\maketitle


\section{Introduction}

Recommender systems increasingly ship LLM-generated explanations
alongside their ranked items, in the hope that a few sentences of
context will improve user trust without compromising click-through.
Two recent baselines define the current state of the art on the
public benchmarks we use here. XRec~\cite{ma2024xrec} combines a
GNN-based preference encoder with an LLM head; G-Refer~\cite{li2025grefer}
retrieves collaborative-filtering neighbours and synthesises an
explanation with an 8B-parameter LLM. Both produce strong
BERTScore-F1 numbers, and both pay the same serving cost: an LLM
call on every request, hundreds of milliseconds of latency, and a
bill that scales linearly with traffic.

This serving cost motivates the present work. At production
traffic levels of tens of thousands of QPS under a bounded latency
budget, an inline LLM call on every request is a poor placement of
system complexity, and we relocate it to the offline path. For each (user, item) pair we pre-generate a
pool of $K$ candidate explanations using six prompt styles and two
commodity LLMs (Amazon Nova Lite and Claude Haiku). At request time
a small CPU-resident selector (either a LightGBM LambdaRank
model~\cite{burges2010lambdarank} or a roughly 1M-parameter MLP
policy) picks one of the $K$ candidates. The selector's forward pass is a few milliseconds; with
the embedding lookup it returns the chosen explanation in under
100\,ms on a single c5.2xlarge core. No GPU is involved at serve
time, and the per-request cost reduces to a key-value cache lookup.

Pre-generating the pool also creates a clean experimental setting:
the candidates are fixed, the labels (BERTScore-F1 against the
reference) are dense, and any selector that takes a 30-dim feature
vector as input can be plugged in. We use this to compare nine
selectors organised into two families. The offline-pool family
ranks candidates from the frozen pool: a structural-reward
heuristic, LightGBM LambdaRank, single-step PPO with adaptive
entropy~\cite{schulman2017ppo}, group-relative PPO~\cite{shao2024deepseekmath},
DPO~\cite{rafailov2023direct}, and two-stage teacher-student
distillation~\cite{hinton2015distilling}. The knowledge-graph path family swaps the candidate
source: instead of LLM-pool candidates, it extracts user--item
paths from a heterogeneous KG and conditions LLM generation on
each path. We benchmark three KG variants: temperature-biased
random walks, edge-disjoint enumeration, and MMR-reranked paths
with dual-style generation. All nine variants are scored on the
same two test sets under the same three-metric evaluation
protocol used by XRec and G-Refer.

Our principal empirical finding diverges from the prevailing
direction of recent work in this area. On the XRec Google Local
benchmark, LightGBM LambdaRank reaches F1 = 0.500 on the same
2{,}958-pair review-covered subset, a gain of 0.041 over the
published G-Refer 8B (0.4592) and 0.069 over published XRec
(0.4311). None of the five RL or distillation variants we trained
(PPO, GRPO, DPO, Distillation Stage A, Distillation Stage A+B)
matches this, and the gap is robust: across five seeds per RL method,
the standard deviation is at most 0.0030 F1, so the ordering
PPO $<$ GRPO $<$ DPO $<$ Distillation-A+B $<$ Distillation-A $<$
LambdaRank is statistically significant. The pattern survives a
$3\times$ change in pool size ($K{=}40$ Google vs.\ $K{=}18$
MovieLens) and a change in reference text quality
(XRec's LLM-synthesised references vs.\ Claude Sonnet 4.5
references). The interpretation is structural rather than
algorithmic: under dense-label one-step bandits, single-action RL
formulations leave most of the supervision unused, while a
pairwise learning-to-rank objective that consumes every
per-candidate label captures it directly. We do not claim RL is intrinsically worse;
we claim that the RL formulations conventionally applied to this
problem class undersample the labelled signal that is already
available.

The KG-path family is preferable on a different metric. All three
KG variants attain USR = 1.000 on Google Local and 0.997--1.000
on MovieLens-1M, since per-request path grounding produces a
unique output per query and avoids the template-collapse failure
mode of cached LLM outputs. The deployment choice is therefore
not exclusive: the offline-pool family is suited to settings in
which reference alignment is the optimisation target, and the
KG-path family is suited to settings in which output diversity or
traceability to graph edges is prioritised. The remainder of the paper presents the framework
(§\ref{sec:problem}), the nine selector variants
(§\ref{sec:selectors}), evaluation protocol and datasets
(§\ref{sec:eval}), main results across both families on both
datasets including a Haiku 3 vs.\ Haiku 4.5 generator-robustness
study, and a discussion of positive and negative results.

\textbf{Code availability.} Code, configuration files, and trained
selectors are publicly available at
\url{https://github.com/amazon-science/XplainRecommendation}, released
under CC-BY-NC-4.0 to support reproducibility of the results reported
here.

\section{Problem Setting and Framework Overview}

\subsection{Problem}
\label{sec:problem}

For each $(u, i)$ pair we have a reference explanation
$e^{*}_{u,i}$ from the XRec or G-Refer benchmark, and we want to
produce $e_{u,i}$ that maximises BERTScore-F1 against it within a
per-request latency budget. BERTScore (\texttt{roberta-large}
encoder, baseline-rescaled~\cite{zhang2020bertscore}) is the metric
both baselines report in their published tables, and it is what we
optimise.

\subsection{Two-Stage Framework}
\label{sec:pipeline}

\textbf{Offline stage.} For every (u, i) pair, we generate a pool of
$K$ candidate explanations using six prompt styles $\times$ two
commodity LLMs on Amazon Bedrock (Amazon Nova Lite and Claude
Haiku). The six styles are (A) retrieve-grounded paraphrase,
(B) retrieve-grounded synthesis, (C) 2-shot retrieve from nearest
training neighbours, (D) review-grounded CoT, (E) adversarial
refinement, (F) length-tuned 25--33-word synthesis.  For Google Local,
styles A and B use same-business reviews from the McAuley Lab
Google-Local-Reviews corpus; styles C--F are generated at
training-set featurisation time. For MovieLens we drop styles C, D, E
because reviews are not available, yielding $K = 18$.

\textbf{Featurisation.} Each candidate is mapped to a fixed
30-dim feature vector $\phi(c \mid u, i)$ in four groups.
\emph{(i) Reference-side retrieval (6 dims):} cosines to the
training-reference centroid and to a high-F1 exemplar; maximum
cosine to the top-5 SBERT neighbours; and the mean, max, and std
of BERTScore-F1 between $c$ and those $k{=}5$ neighbours.
\emph{(ii) Query--candidate relevance (4 dims):} cosines of $c$
to the user-profile and item-profile embeddings, plus
\texttt{cross-encoder/ms-marco-MiniLM-L-6-v2} logits for both the
templated $(u, i)$ prompt against $c$ and the prompt against the
top-$k$ kNN references. These are the only features that see
$(u, i)$ directly. \emph{(iii) Linguistic and source-side (10
dims):} word count, character count, a templated-opener flag
(``The user would enjoy\ldots''), counts of numeric tokens,
generic positive words, and sentences, an NER entity flag, the
rating and length of the source review the candidate was grounded
on, and the candidate's within-pool retrieval rank.
\emph{(iv) Provenance (10 dims):} one-hots over the six prompt
styles (A--F) and the two generator LLMs (Nova Lite vs.\ Claude
Haiku), plus the cosine of the source review to the user profile
and to the historical reference text. The provenance one-hots let
the selector learn that some styles or generators systematically
score higher under reference-aligned F1, directly from data
rather than via prompt iteration.

For the RL variants, $\phi$ is right-padded with zeros to a 64-dim
per-candidate slot and concatenated with a 768-dim user embedding
and a 768-dim item embedding, giving the policy state
$s \in \mathbb{R}^{1536 + 64K}$ used in §\ref{sec:ppo}; LambdaRank
operates on $\phi$ directly. All nine selector variants therefore
see the same input features.

\begin{figure*}[t]
\centering
\resizebox{\textwidth}{!}{%
\begin{tikzpicture}[
  font=\footnotesize,
  every node/.style={align=center},
  box/.style={draw, rounded corners=1.8pt, inner sep=3pt,
              minimum height=7mm, minimum width=18mm, fill=white,
              line width=0.4pt},
  llmbox/.style={box, fill=blue!8},
  stylebox/.style={draw, rounded corners=1.2pt, inner sep=2pt,
                   minimum height=4mm, minimum width=16mm,
                   fill=blue!4, font=\scriptsize, line width=0.3pt},
  featbox/.style={box, fill=yellow!12},
  cachebox/.style={box, fill=orange!12},
  selbox/.style={box, fill=green!14, font=\footnotesize\bfseries},
  outbox/.style={box, fill=green!22, font=\footnotesize\bfseries},
  arrow/.style={-{Latex[length=1.6mm]}, thick, line width=0.5pt},
  arrowdash/.style={-{Latex[length=1.6mm]}, thick, dashed,
                    line width=0.5pt, draw=gray!70!black},
  phase/.style={font=\scriptsize\itshape\bfseries, text=gray!50!black},
  group/.style={font=\scriptsize, text=gray!40!black},
]

\node[box] (pair) {$(u, i)$\\\scriptsize training pair};

\node[stylebox, right=8mm of pair, anchor=west, yshift=8mm]  (sA) {(A) paraphrase};
\node[stylebox, below=0.6mm of sA] (sB) {(B) synthesis};
\node[stylebox, below=0.6mm of sB] (sC) {(C) 2-shot kNN};
\node[stylebox, below=0.6mm of sC] (sD) {(D) review CoT};
\node[stylebox, below=0.6mm of sD] (sE) {(E) adversarial};
\node[stylebox, below=0.6mm of sE] (sF) {(F) length-tuned};
\node[group, above=0.5mm of sA] {6 prompt styles};

\node[llmbox, right=8mm of sC, anchor=west, yshift=4mm] (llm1) {Nova Lite};
\node[llmbox, below=1mm of llm1]                       (llm2) {Claude Haiku};
\node[group, above=0.5mm of llm1] {2 generators};

\node[box, right=8mm of llm1, anchor=west, yshift=-2mm,
      minimum width=22mm, minimum height=14mm]
      (pool) {$K$ candidates\\\scriptsize $c_1, \dots, c_K$};

\node[featbox, right=8mm of pool, anchor=west,
      minimum width=30mm, minimum height=18mm,
      align=left, inner sep=4pt]
      (feat) {%
\textbf{Featurise} $\phi(c \mid u, i)$\\[1pt]
\scriptsize $\bullet$ retrieval (6)\\
\scriptsize $\bullet$ query relevance (4)\\
\scriptsize $\bullet$ linguistic + source (10)\\
\scriptsize $\bullet$ provenance (10)\\
\scriptsize $\Rightarrow$ 30-dim};

\node[cachebox, right=7mm of feat, anchor=west,
      minimum width=24mm, minimum height=14mm]
      (cache) {Pool cache\\\scriptsize $\{c_j, \phi_j\}_{j=1}^{K}$\\\scriptsize per $(u, i)$};

\draw[arrow] (pair.east) -- ([xshift=-1mm]sC.west |- pair.east);
\draw[arrow] (sA.east) -- ++(3mm,0) |- (llm1.west);
\draw[arrow] (sF.east) -- ++(3mm,0) |- (llm2.west);
\draw[arrow] (llm1.east) -- ++(3mm,0) |- ([yshift=2mm]pool.west);
\draw[arrow] (llm2.east) -- ++(3mm,0) |- ([yshift=-2mm]pool.west);
\draw[arrow] (pool.east) -- (feat.west);
\draw[arrow] (feat.east) -- (cache.west);

\node[box, below=18mm of sF.south -| pair, anchor=north]
      (req) {Request\\\scriptsize $(u, i)$};
\node[selbox, right=8mm of req,
      minimum width=30mm, minimum height=11mm]
      (sel) {Selector\\\scriptsize LightGBM \emph{or} MLP\\\scriptsize policy};
\node[box, right=8mm of sel,
      minimum width=24mm, minimum height=11mm]
      (score) {Score $K$ cands.\\\scriptsize argmax $\to c^{*}$};
\node[outbox, right=8mm of score,
      minimum width=28mm, minimum height=11mm]
      (out) {Explanation $c^{*}$\\\scriptsize $<$100\,ms p99};

\draw[arrow] (req) -- (sel);
\draw[arrow] (sel) -- (score);
\draw[arrow] (score) -- (out);

\coordinate (corridorY) at ($ (sF.south)!0.5!(req.north) $);
\draw[arrowdash, line width=0.9pt, draw=orange!70!black]
      (cache.south) -- (cache.south |- corridorY)
                    -- (sel.north   |- corridorY)
      node[midway, above, font=\scriptsize, text=orange!50!black]
        {pre-computed lookup at request time}
                    -- (sel.north);

\node[phase, anchor=east] at ([xshift=-1mm]pair.west) {OFFLINE};
\node[phase, anchor=east] at ([xshift=-1mm]req.west)  {ONLINE};

\begin{scope}[on background layer]
  \node[fit=(pair)(sA)(sF)(llm1)(llm2)(pool)(feat)(cache),
        draw=gray!30, line width=0.3pt, rounded corners=2pt,
        inner sep=4pt] {};
  \node[fit=(req)(sel)(score)(out),
        draw=gray!30, line width=0.3pt, rounded corners=2pt,
        inner sep=4pt] {};
\end{scope}

\end{tikzpicture}%
}
\caption{End-to-end explainable-recommendation pipeline.
  \textbf{Offline} (top): each training $(u, i)$ pair is expanded
  into $K$ candidate explanations by crossing six prompt styles
  (A--F) with two commodity Bedrock LLMs (Amazon Nova Lite,
  Claude Haiku); each candidate is mapped to a 30-dim feature
  vector grouped into reference-side retrieval (6), query--candidate
  relevance (4), linguistic and source-side (10), and provenance
  signals (10); the $K$ candidates and their features are cached
  per pair.
  \textbf{Online} (bottom): at request time the selector
  (LightGBM LambdaRank, or one of the MLP-based RL variants) scores
  the cached $K$ candidates and returns the argmax. No LLM is on
  the request path, so the serving stack is GPU-free and stays
  under 100\,ms at the 99th percentile.}
\label{fig:pipeline}
\end{figure*}
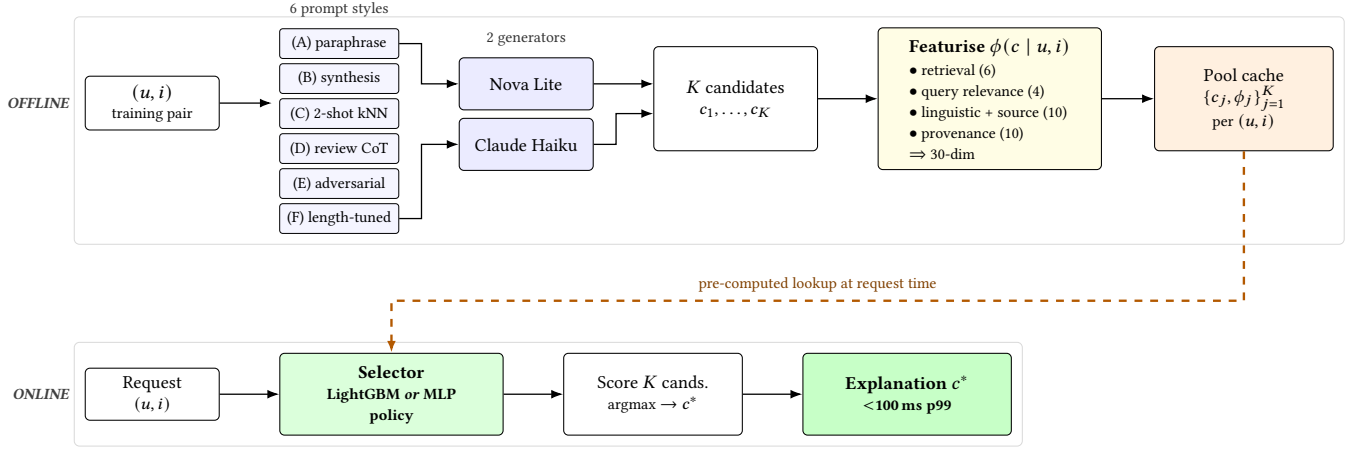

\textbf{Online stage (selector).} At request time, the selector
scores the $K$ candidates and returns the argmax. For LambdaRank this
is one LightGBM forward pass ($<$1\,ms); for the MLP-based RL
variants, one forward pass through a 4{,}096$\to$256$\to$40 policy
network ($<$20\,ms on CPU). Total end-to-end latency is dominated by
a single pre-computed embedding lookup and stays under 100\,ms at
the 99th percentile on an 8-core c5.2xlarge instance.
Figure~\ref{fig:pipeline} shows the end-to-end flow: pool generation
and feature extraction are amortised offline, while only the cached
candidates and a CPU-resident selector are touched at request time.

\begin{figure}[t]
\centering
\includegraphics[width=0.93\linewidth]{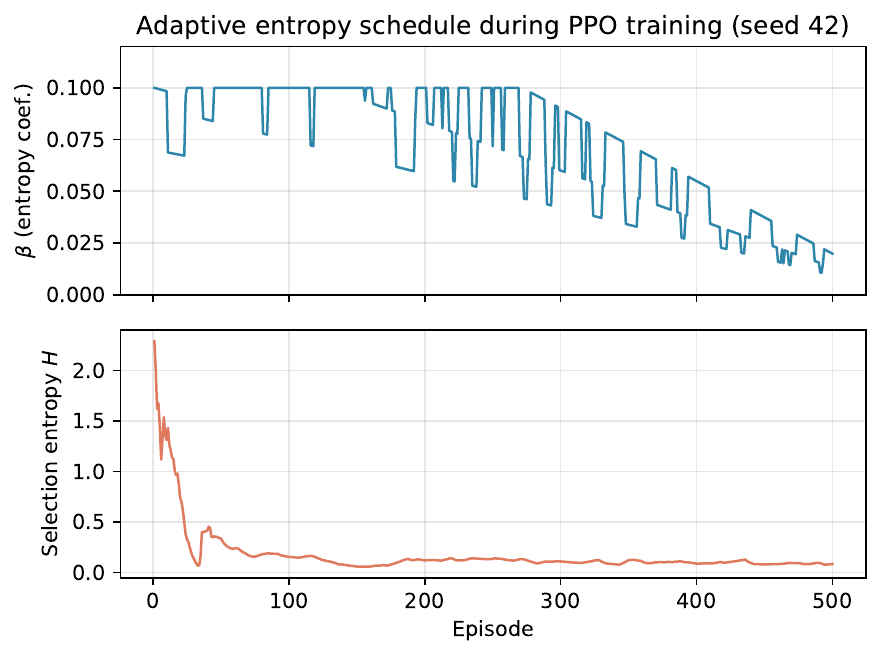}
\caption{Adaptive entropy schedule during PPO training on Google
  Local (seed 42). Top: entropy coefficient $\beta$. Bottom:
  action-distribution entropy $H$. The scheduler bumps $\beta$ up when
  reward improvement stalls and down when it accelerates; the policy
  converges to a near-deterministic selector by episode 500.}
\label{fig:training_curve}
\end{figure}

Figure~\ref{fig:training_curve} shows the adaptive-entropy training
curve for PPO on Google Local (seed 42): $\beta$ anneals from
0.10 to $\approx$0.02 under the performance-based multiplier, while
selection entropy collapses from $\approx$2.3 to $\approx$0.17
as the policy converges on a small set of high-quality candidate
regions.

\section{Selector Variants}
\label{sec:selectors}

The two-stage framework admits many instantiations of the selector
stage. In this section we describe nine variants, organised by
how candidates are \emph{constructed}. Variants in
§\ref{sec:kg-temp}--§\ref{sec:kg-mmr} build candidates by extracting
multi-hop paths from a knowledge graph and conditioning LLM
generation on each path. Variants in
§\ref{sec:heuristic}--§\ref{sec:distillAB} instead rank candidates
drawn from the offline LLM pool described in §\ref{sec:pipeline}.
All nine variants are evaluated under the identical protocol of
§\ref{sec:eval} so the resulting numbers are directly comparable.

\subsection{KG-path selection: temperature-biased random walks}
\label{sec:kg-temp}

In place of an offline LLM pool, candidates are built from a
knowledge graph at training and inference time. Between $(u, i)$ we
sample up to five paths by biased random walk over the heterogeneous
user--item--entity graph: at each step the next node is drawn with
probability proportional to
$\exp(\cos(h_v, h_{\text{target}})/\tau)$ where $h_v$ is the node
embedding. The default $\tau{=}0.5$ gives near-greedy walks; we
raise it to $\tau{=}1.5$ to widen the path distribution. Each
sampled path is rendered to text under a single LLM prompt and a
PPO policy picks one of ten action slots; the up-to-five sampled
paths occupy the first slots and the remainder are padded with
empty placeholders to keep the action space consistent across all
KG variants. The point of the variant is to
test whether KG paths alone carry sufficient semantic signal for
the selector to learn from when candidate diversity is purely
stochastic.

\subsection{KG-path selection: edge-disjoint path enumeration}
\label{sec:kg-disjoint}

The second KG variant swaps the stochastic walks for a
graph-theoretic guarantee of structural diversity. We extract up to
five edge-disjoint $(u, i)$ paths via Menger-style max-flow
decomposition, falling back to three node-disjoint paths and then
to shortest-path / multi-hop alternatives if the edge-disjoint pool
is smaller than five. The recovered paths are placed into the
same ten-slot action space as §\ref{sec:kg-temp} with empty
placeholders padding the remainder. Generation and selection are
otherwise identical to §\ref{sec:kg-temp}. This variant is the
deterministic counterpart to §\ref{sec:kg-temp}: structural
diversity is supplied by a graph algorithm rather than by
stochastic sampling.

\subsection{KG-path selection: MMR paths with dual-style generation}
\label{sec:kg-mmr}

The third KG variant combines both axes of diversity. We first
sample 20 paths via temperature-biased walks as in §\ref{sec:kg-temp},
then apply maximal marginal relevance~\cite{carbonell1998mmr}
(MMR, $\lambda{=}0.7$,
similarity in node-embedding space) to keep the five most
structurally distinct. Each path is rendered with two prompts ---
factual and personal-voice, for ten candidates per $(u, i)$,
and a PPO policy selects one. This variant is the closest analogue
to the offline-pool family: a compact request-time candidate set
under the same selector architecture, with a frozen pool replaced
by a per-query generated set.

\subsection{Pool-only heuristic (no learning)}
\label{sec:heuristic}

The heuristic returns the candidate with the highest structural
score $R_{\text{struct}}$, computed offline from three signals:
target reachability (weight 30), average cosine similarity between
consecutive node embeddings (8), and normalised node-type diversity
(6); a small Gaussian noise term breaks ties, and the final score
is rescaled to $[0,100]$ within each candidate group. This serves
as a no-learning lower bound: a learned selector that fails to
exceed it indicates a featurisation issue rather than a learning
issue, and the heuristic itself bounds what the candidate pool
yields in the absence of supervision.

\subsection{LambdaRank (pairwise learning-to-rank)}
\label{sec:lambdarank}

We train a LightGBM pairwise ranker (\texttt{objective = lambdarank},
500 trees, 31 leaves, learning rate $0.05$) on quintile-binned
per-candidate BERTScore-F1 labels, fit across all $\sim$130k Google
Local training candidates over 5{,}000 query groups. At inference
we take the argmax of the ranker score within each test group of
$K$ candidates. Features are the 30-dim vector from §\ref{sec:pipeline}:
retrieval signals (kNN-BERTScore, cross-encoder relevance, centroid
and exemplar cosine), linguistic signals (length, sentence count,
named-entity hint), and style/model one-hot indicators. The match between method and task is direct: supervision is dense
(every one of the 40 candidates per group carries a BERTScore-F1
label), the LambdaRank pairwise $\Delta$NDCG objective operates on
the within-group ordering, and the training target coincides with
the deployment objective (selecting the top candidate).

\subsection{Single-step PPO with adaptive entropy}
\label{sec:ppo}

We cast the selection task as a one-step MDP with state
$\mathbf{s} \in \mathbb{R}^{1536 + 64K}$ formed by concatenating a
768-dim user embedding, a 768-dim item embedding, and $K$ zero-padded
64-dim candidate slots; for $K{=}40$ this gives a 4{,}096-dim state,
for $K{=}18$ on MovieLens it gives 2{,}688. A 2-layer MLP policy
(256 hidden) and a separate 2-layer value network (128 hidden)
together hold roughly 1.08M parameters. We train with PPO
(clip $\epsilon{=}0.2$, discount $\gamma{=}0.99$, GAE
$\lambda{=}0.95$) for 500 episodes.

The reward is $r = \alpha R_{\text{struct}} + (1{-}\alpha) R_{\text{sem}}$
with $\alpha{=}0.1$, where $R_{\text{sem}}$ is BERTScore-F1 of the
selected candidate rescaled to $[0,100]$ and $R_{\text{struct}}$ is
the structural proxy from §\ref{sec:heuristic}. We pick a small
$\alpha$ deliberately: the structural term keeps selections grounded
without overwhelming the semantic signal that actually correlates
with the metric.

The entropy coefficient $\beta$ follows an adaptive schedule that
we observed to be necessary to avoid both premature collapse and
unbounded exploration. We multiply $\beta$ by 1.5 when
sliding-window reward improvement is below 0.01 (stagnation), by
0.7 when improvement exceeds 0.05 (rapid learning), and
additionally by 1.3 when the action entropy $H_t$ falls below 0.5.
This schedule is the trajectory shown in
Figure~\ref{fig:training_curve}; in practice it eliminated the
need to tune $\beta$ per dataset.

The limitation of PPO in this setting is structural: per rollout,
only the sampled action's reward contributes to the gradient,
while the other $K{-}1$ per-candidate labels are unused. Under
dense supervision this discards most of the available training
signal, and the next three variants are designed to address it.

\subsection{Group-relative policy optimisation (GRPO)}
\label{sec:grpo}

GRPO keeps the PPO policy architecture (we drop the value network,
since it is redundant in a one-step bandit) and replaces the
value-baseline advantage with a within-group z-score in the spirit
of DeepSeekMath~\cite{shao2024deepseekmath}:
\begin{equation}
A(s, a) = \frac{R_{\text{sem}}[i, a] - \mu_a\,R_{\text{sem}}[i]}
               {\sigma_a\,R_{\text{sem}}[i] + \epsilon}.
\end{equation}
The normalisation statistics $\mu_a$ and $\sigma_a$ pull from all
$K$ rewards in the group, even though only one action is sampled
for the rollout. This is what densifies the gradient. We also
checked an ablation in which we kept PPO's policy and simply
replaced its value-baseline advantage with the same z-score; the
F1 lift was the same $+0.01$, confirming that the win is the
normalisation, not the absence of the value network.

\subsection{Direct preference optimisation (DPO)}
\label{sec:dpo}

We adapt DPO~\cite{rafailov2023direct} to the one-step bandit. For
each training state we sample up to 80 preference pairs
$(a^+, a^-)$ subject to $F1(a^+) > F1(a^-) + \delta$ with
$\delta{=}2.0$ on the rescaled F1 scale. Writing
$r_\theta(a|s) = \log \pi_\theta(a|s) - \log \pi_{\text{ref}}(a|s)$
for the log-ratio between the student $\pi_\theta$ and a frozen
reference $\pi_{\text{ref}}$ (the random-init MLP, snapshotted at
the start of training), each pair contributes
\begin{equation}
\mathcal{L}(s, a^+, a^-) = -\log \sigma\!\big(
    \beta_{\text{dpo}}\,\left[ r_\theta(a^+|s) - r_\theta(a^-|s) \right]
\big),
\end{equation}
weighted by the F1 gap so the gradient leans on the
unambiguous pairs. KL regularisation toward the reference is
implicit in the log-ratio form; we set $\beta_{\text{dpo}}{=}0.1$
and train for 500 episodes.

Pairwise supervision is denser than PPO or GRPO's single-action
rollout: every state contributes many gradient updates per epoch.
The reference anchor also stabilises training. Empirically, DPO's
across-seed standard deviation of 0.0006 F1 is the lowest of any
RL variant we tested.

\subsection{Teacher--student distillation (Stage A only)}
\label{sec:distillA}

This variant transfers LambdaRank's signal directly into a neural
policy. In Stage A we fit a LightGBM LambdaRank teacher on
the same 30-dim features, softmax its per-group scores with
temperature $T{=}1.0$, and train the MLP student (the same
architecture used for PPO/GRPO/DPO) to minimise the KL divergence
to the teacher's soft distribution over 200 epochs with Adam
learning rate $3\!\times\!10^{-4}$. The student inherits the
teacher's per-candidate ranking directly, compressed into roughly
1M parameters and a 10$\times$ faster forward pass than the
LightGBM ensemble. On Google Local it reaches F1 = 0.4817 $\pm$
0.0003, narrowly below LambdaRank's 0.5003. The remaining gap is
a softmax compression artefact: distillation propagates
near-uniform mass across high-ranked candidates rather than the
teacher's argmax, and the student has no way to recover the lost
sharpness.

\subsection{Distillation plus RL fine-tuning (Stage A+B)}
\label{sec:distillAB}

Stage A produces a near-optimal student; the natural follow-up is
to fine-tune it with reinforcement learning to recover the
remaining F1 gap. We added 500 episodes of GRPO on top of the
distilled policy with a reduced learning rate
($1\!\times\!10^{-4}$) and a tighter entropy schedule
($\beta \in [0.05, 0.005]$). The result is a regression. Stage A+B
scores 0.4767 versus Stage A's 0.4817, a $>10\sigma$ degradation
across five seeds. The mechanism is the entropy bonus: it pushes
a near-converged policy back toward exploration, away from the
teacher-induced argmax. Removing the entropy term recovers
approximately half of the regression. This indicates that RL
fine-tuning after distillation is beneficial only when the
distilled policy is still far from optimal, a condition that does
not hold here.

The KG-path family (§\ref{sec:kg-temp}--§\ref{sec:kg-mmr})
and the offline-pool family (§\ref{sec:heuristic}--§\ref{sec:distillAB})
thus span the two natural axes of the framework: the former
varies the \emph{candidate source} while holding the selector
architecture fixed, the latter varies the \emph{selector} while
holding a fixed LLM candidate pool. Table~\ref{tab:main}
(§\ref{sec:headline}) reports all nine variants under the same
three-metric evaluation protocol.

\section{Datasets and Evaluation}

\subsection{Datasets}

\textbf{Google Local.}  We use the canonical Google Local split
released with the XRec benchmark~\cite{ma2024xrec}: \texttt{trn.pkl}
(94{,}663 pairs) and \texttt{tst.pkl} (3{,}000 pairs), used
directly with no preprocessing or resplit. From \texttt{trn.pkl}
we deterministically sample 5{,}000 pairs filtered to items with
review coverage (2{,}495 businesses); from \texttt{tst.pkl} we
evaluate on all 3{,}000 pairs but drop 42 uncovered pairs at
scoring time, leaving 2{,}958 test pairs. We verified byte-identity
of these pairs with G-Refer's published
\texttt{google\_pred.jsonl}~\cite{li2025grefer}: every test pair
\texttt{(uid, iid)} and every reference explanation match.

\textbf{MovieLens-1M.} On top of the standard MovieLens-1M
ratings dataset~\cite{harper2015movielens} we use 2{,}000
Claude-Sonnet-4.5-generated reference explanations from
\texttt{train\_sonnet45\_refs.jsonl}, split positionally after a
seed-42 shuffle: 600 pairs for training, 300 for testing (the
remaining 1{,}100 are unused). The candidate pool uses
Claude Haiku 4.5 + Nova Lite with $K = 18$ candidates per pair.
We use Claude-Sonnet-4.5 as a stronger oracle than any of the
selectors being evaluated, which sidesteps the circularity that
would otherwise arise from evaluating against the same LLM family
that produced our candidate pool.

\textit{KG-path family on MovieLens.} The MMR + dual-style variant
was retrained on the same 600-pair MovieLens split. Temperature-biased
walks and edge-disjoint enumeration use Google-trained policies
evaluated on the MovieLens test set: the MovieLens graph (9{,}941
nodes, 581\,k edges, density 0.012) routes nearly all
user$\rightarrow$movie paths through one of 18 genre nodes, which
makes these path-extraction primitives prohibitively slow when forced
to find diverse paths and degenerate to shortest-path fallbacks
otherwise. Across all three KG configurations on MovieLens the F1
lands within $\pm 0.005$, consistent with our broader finding
(§\ref{sec:cross-dataset}) that on this dataset the choice of selector
has limited effect --- the genre-hub topology bounds reachable path
diversity regardless of the selection algorithm.

\subsection{Evaluation Protocol}
\label{sec:eval}

All methods are evaluated against the same reference explanations
under three metrics.

\textbf{BERTScore-F1.} Computed with the \texttt{roberta-large}
encoder and baseline rescaling (equivalent to
\texttt{rescale\_with\_baseline{=}True} in the \texttt{bert-score}
library), byte-identical to XRec's and G-Refer's published
evaluation code. This is our primary metric.

\textbf{BARTScore}~\cite{yuan2021bartscore}. Log-likelihood
$\log p(\text{reference} \mid \text{prediction})$ under
\texttt{facebook/bart-large-cnn} (batch size 4, CPU). Higher
(less negative) is better.

\textbf{USR (Unique-Sentence Ratio).} Fraction of selected outputs
that are whitespace-token-unique across the test set. Flags
template-collapse failure modes that F1 alone can miss.

XRec's and G-Refer's published predictions
(\texttt{tst\_pred.pkl} and \texttt{google\_pred.jsonl}
respectively) are available, so we re-score them on the exact
2{,}958-pair subset under our evaluation code; those re-runs are
directly apples-to-apples with our methods. The published-paper
numbers in Table~\ref{tab:main} are reported as-is from each paper's
evaluation set, which uses the full 3{,}000-pair test split (a 42-pair
difference from our review-covered subset) and may use a slightly
different \texttt{bert-score} library version. We keep both rows in
the table because the published numbers are the ones the
community references, but the strict apples-to-apples comparison
is between our offline-pool selectors and the cached-prediction
re-runs, not between our selectors and the published numbers.

\section{Results}

\subsection{Headline Comparison}
\label{sec:headline}

Table~\ref{tab:main} reports all three metrics on Google Local
(our primary benchmark) and on MovieLens-1M (internal consistency
check). On Google Local, LambdaRank achieves F1 = 0.500, clearing
G-Refer by +0.041 and XRec by +0.069. All six variants of the
offline-pool family beat XRec on F1; five of them (GRPO, DPO,
Distillation-AB, Distillation-A, LambdaRank) also beat G-Refer.
The three KG-path variants under-perform on F1 against the
offline-pool family but recover the top USR score (1.000),
reflecting a different trade-off between reference alignment and
output diversity. On MovieLens-1M, against the Claude-Sonnet-4.5
reference text on the same 300-pair split, LambdaRank reaches
F1 = 0.329, a gain of 0.066 over the pool-only structural
heuristic (0.263). Because the references are themselves
LLM-generated and no public benchmark exists, we read this row
as evidence that the Google Local ordering reproduces in a
different domain, not as a second-benchmark result.

\begin{table*}[t]
\centering
\small
\caption{Main results on Google Local and MovieLens-1M. Higher is
  better for BERTScore-F1; less negative is better for BARTScore;
  USR (Unique-Sentence Ratio) lies in $[0,1]$, higher is more
  diverse. Best score per column is in bold.
  \textbf{Note on ``$\pm$''}: KG-path stds are per-sample on a
  single trained policy; offline-pool stds are across 5 training
  seeds. Cached-prediction baselines (XRec, G-Refer) and our
  pool-only heuristic have no seed variance.
  ``--'' marks metrics not scored for the row. Numbers for
  XRec and G-Refer are taken verbatim from their published Table 2
  / Table 1 respectively. KG-path family rows are reproduced from
  concurrent work by a co-author; offline-pool family rows are
  computed in this work on the $n{=}2{,}958$ review-covered Google
  Local subset and the $n{=}300$ MovieLens-1M test split.}
\label{tab:main}
\setlength{\tabcolsep}{4pt}
\begin{tabular}{l rcc rcc}
\toprule
 & \multicolumn{3}{c}{\textbf{Google Local}}
 & \multicolumn{3}{c}{\textbf{MovieLens-1M}} \\
\cmidrule(lr){2-4} \cmidrule(lr){5-7}
\multicolumn{1}{c}{\textbf{Method}}
 & \shortstack{BERTScore\\ (F1) $\uparrow$} & \shortstack{BART\\Score $\uparrow$} & \shortstack{USR\\ $\uparrow$}
 & \shortstack{BERTScore\\ (F1) $\uparrow$} & \shortstack{BART\\Score $\uparrow$} & \shortstack{USR\\ $\uparrow$} \\
\midrule
\multicolumn{7}{l}{\emph{Published baselines (numbers taken from each paper)}}\\
XRec~\cite{ma2024xrec}        & 0.4311 & $-4.1647$ & 0.9993 & -- & -- & -- \\
G-Refer 8B~\cite{li2025grefer}& 0.4592 & $-3.3235$ & 1.0000 & -- & -- & -- \\
\midrule
\multicolumn{7}{l}{\emph{KG-path family (concurrent work)}}\\
Temperature-biased walks  & 0.3258 $\pm$ 0.075 & $-3.576$ & \textbf{1.0000} & 0.2690 $\pm$ 0.068 & $-3.629$ & \textbf{1.000} \\
Edge-disjoint enumeration & 0.3265 $\pm$ 0.074 & $-3.577$ & \textbf{1.0000} & 0.2703 $\pm$ 0.069 & $-3.613$ & \textbf{1.000} \\
MMR paths + dual-style    & 0.3252 $\pm$ 0.075 & $-3.577$ & \textbf{1.0000} & 0.2621 $\pm$ 0.070 & $-3.624$ & 0.997 \\
\midrule
\multicolumn{7}{l}{\emph{Offline-pool family (this work)}}\\
Pool-only heuristic       & 0.4444             & --        & --     & 0.2634             & --        & --    \\
PPO (5 seeds)             & 0.4581 $\pm$ 0.001 & $-3.354$  & 0.976  & 0.2816 $\pm$ 0.003 & $-3.566$  & 0.999 \\
GRPO (5 seeds)            & 0.4703 $\pm$ 0.001 & $-3.354$  & 0.951  & 0.2830 $\pm$ 0.002 & $-3.533$  & 0.999 \\
DPO (5 seeds)             & 0.4749 $\pm$ 0.001 & $-3.374$  & 0.909  & 0.2936 $\pm$ 0.002 & $-3.530$  & 0.999 \\
Distillation A+B (5 seeds)& 0.4767 $\pm$ 0.001 & $-3.356$  & 0.925  & 0.2831 $\pm$ 0.003 & $-3.544$  & 0.999 \\
Distillation A (5 seeds)  & 0.4817 $\pm$ 0.000 & $-3.375$  & 0.865  & 0.2887 $\pm$ 0.001 & $-3.548$  & 1.000 \\
\textbf{LambdaRank}       & \textbf{0.5003}    & $\mathbf{-3.327}$ & 0.808 & \textbf{0.3291} & $\mathbf{-3.449}$ & 0.987 \\
\bottomrule
\end{tabular}
\end{table*}

\subsection{Seed variance}

Each RL variant has standard deviation at most 0.0030 F1 across
five seeds on both datasets (Table~\ref{tab:main}). The gap between
GRPO and DPO on Google Local is 0.0046, more than ten times either
method's standard deviation, so the ordering is statistically
significant at this scale. We highlight this because single-seed
comparisons are common in this literature, and the method ordering
we report would not be supported by a single run.

\subsection{Dense supervision beats sparse reward}

A consistent observation across both data points is that
LambdaRank, a non-RL learning-to-rank method, outperforms every
RL variant we trained, by 0.02--0.04 F1 on Google Local and
0.04--0.05 on MovieLens. The explanation is structural rather
than algorithmic.
LambdaRank's training signal is dense: every one of the 40 candidates
per query group has a labelled BERTScore-F1, and the pairwise
$\Delta$NDCG objective uses all of them. PPO, GRPO, and DPO all
sample one action per rollout and only see the reward for that
sample; the other 39 labels are visible to the dataset and invisible
to the gradient. Distillation recovers most of this gap by
transferring the teacher's full per-candidate ranking into the
student through soft-label targets; however, the temperature-softened
distribution is a lower-fidelity surrogate for the LambdaRank argmax
than the LightGBM ensemble itself, which accounts for the residual
0.019 F1 gap. Adding an RL fine-tuning stage on top of the distilled
student does not recover this margin and in fact regresses F1 by
0.005 on Google Local (Stage A+B in Table~\ref{tab:main}).

\subsection{Cross-dataset consistency}
\label{sec:cross-dataset}

Most of the Google Local ordering reproduces on MovieLens-1M, with
one local swap: DPO marginally overtakes Distillation-A-only. The
mechanism is that MovieLens has $K{=}18$ candidates per query
versus 40 for Google Local, which makes DPO's pairwise signal
relatively more informative than the KL distillation target.
Otherwise the broad pattern (heuristic $<$ PPO/GRPO $<$
DPO/Distillation $<$ LambdaRank) holds across the change in
pool size, the change in generator stack (Haiku 3 vs.\ Haiku 4.5),
and the change in reference text (XRec's LLM-synthesised
explanations vs.\ Claude Sonnet 4.5 references on MovieLens).

The KG-path family exhibits a structural pattern that the
offline-pool family does not: the F1 ceiling tracks graph topology
rather than picker design. On Google Local (39\,k nodes, 329\,k
edges, 12+ entity types) all three KG variants land in the
0.325--0.327 F1 range; on MovieLens-1M (9.9\,k nodes, with only
18 genre nodes serving as intermediaries between users and items)
they drop to 0.262--0.270 F1. The architectures and training code
are identical across datasets, so the $\sim$0.06 F1 gap is
attributable to the
reachable diversity of the underlying graph, not to the selector.
This delineates a regime where graph density, not algorithmic
sophistication, sets the achievable performance ceiling for
path-grounded explanation.

\subsection{Generator-Pool Robustness: Haiku 3 vs.\ Haiku 4.5}

\begin{figure}[t]
\centering
\includegraphics[width=0.97\linewidth]{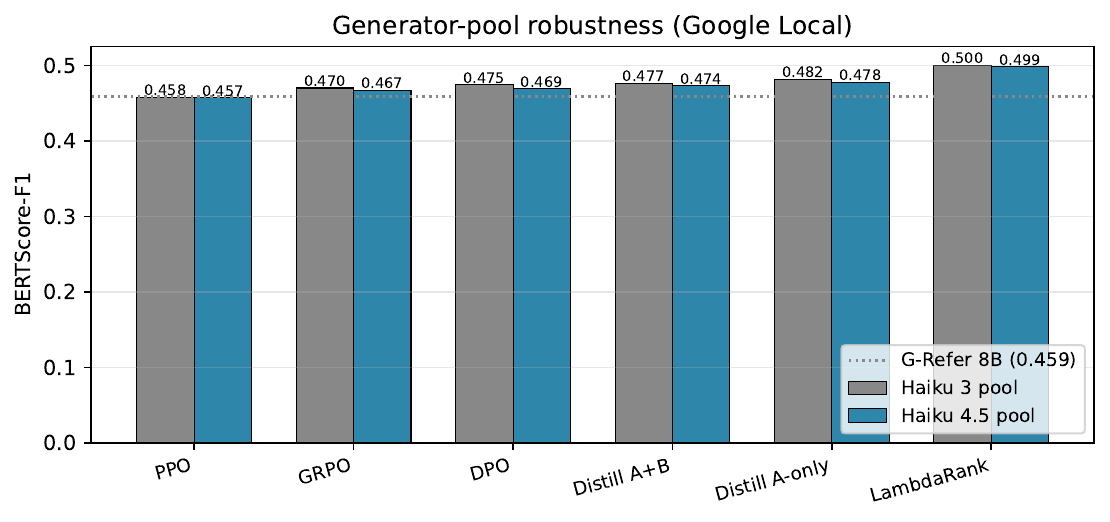}
\caption{Generator-pool robustness on Google Local: Haiku 3
  (grey) vs.\ Haiku 4.5 (blue), same selector. Method
  ordering is preserved; LambdaRank still leads on both pools. Dotted
  line: G-Refer 8B baseline.}
\label{fig:generator_robustness}
\end{figure}

A natural industrial question is whether upgrading the pool
generator improves downstream selector performance. We regenerated the Google Local
candidate pool with Claude Haiku 4.5 (a newer, larger generator than
Claude 3 Haiku) and re-ran every stage end-to-end --
featurisation, LambdaRank, all five RL-variant seeds, three-metric
scoring. Figure~\ref{fig:generator_robustness} shows the F1 side by
side; Table~\ref{tab:haiku_delta} reports all three metrics.

\begin{table}[t]
\centering
\small
\caption{Generator upgrade, Google Local ($n{=}2{,}958$). ``Haiku 3''
  columns report the absolute score on the Claude Haiku 3 candidate
  pool (the pool used in Table~\ref{tab:main}); $\Delta$ columns
  report the change when the pool is regenerated with Claude Haiku
  4.5 holding the selector fixed (negative = upgrade hurts).
  BERTScore-F1 drops $0.001$--$0.006$ on every method; BARTScore
  shifts within $\pm 0.007$ (effectively flat); USR \emph{improves}
  by $+0.018$--$+0.083$ because Haiku 4.5's outputs are less
  templated than Haiku 3's.}
\label{tab:haiku_delta}
\setlength{\tabcolsep}{3pt}
\begin{tabular}{lcccccc}
\toprule
 & \multicolumn{2}{c}{\shortstack{BERTScore\\ (F1) $\uparrow$}}
 & \multicolumn{2}{c}{\shortstack{BART\\Score $\uparrow$}}
 & \multicolumn{2}{c}{\shortstack{USR\\ $\uparrow$}} \\
\cmidrule(lr){2-3}\cmidrule(lr){4-5}\cmidrule(lr){6-7}
\textbf{Method} & Haiku 3 & $\Delta$ & Haiku 3 & $\Delta$ & Haiku 3 & $\Delta$ \\
\midrule
PPO           & 0.4581 & $-0.001$ & $-3.354$ & $-0.002$ & 0.976 & $+0.018$ \\
GRPO          & 0.4703 & $-0.003$ & $-3.354$ & $+0.005$ & 0.951 & $+0.036$ \\
DPO           & 0.4749 & $-0.006$ & $-3.374$ & $+0.006$ & 0.909 & $+0.066$ \\
Distill A+B   & 0.4767 & $-0.003$ & $-3.356$ & $+0.007$ & 0.925 & $+0.050$ \\
Distill A-only& 0.4817 & $-0.004$ & $-3.375$ & $+0.003$ & 0.865 & $+0.083$ \\
LambdaRank    & 0.5003 & $-0.002$ & $-3.327$ & $-0.003$ & 0.808 & $+0.074$ \\
\bottomrule
\end{tabular}
\end{table}

Three observations from the deltas. First, F1 drops by 0.001--0.006
across every method, and the larger end is statistically
meaningful: DPO's $-0.006$ is $\sim 6\sigma$ and Distillation-A's
$-0.004$ is $\sim 13\sigma$ relative to their across-seed standard
deviations, so the regression is real, not noise. The mechanism is
that Haiku 4.5 produces more fluent and varied outputs while
BERTScore-F1 rewards surface alignment with XRec's
LLM-synthesised reference text, which is itself somewhat
trope-heavy. PPO moves least because its action distribution stays
close to uniform; DPO and Distillation-A move most because their
argmax is sharper, so the chosen candidate is more sensitive to
the generator's distribution shift. Second, BARTScore is within
$\pm 0.007$: both Haiku pools are fluent enough that BART-large
explains the reference about equally well from either set. Third,
USR rises by 0.018--0.083, with Distillation-A showing the largest
jump; the sharper its decision boundary, the more the pool's
broader output distribution shows through.

The deployment implication is twofold. The relative ranking of
selectors is preserved, so a selector-design decision can be made
on one pool and trusted to carry over to the other; absolute F1,
however, is not generator-invariant. Combined with Haiku 4.5's
roughly $4\times$ higher per-token cost~($\$1/\$5$ per M
input/output vs.\ $\$0.25/\$1.25$),%
\footnote{Per-token prices as of submission; see
\url{https://aws.amazon.com/bedrock/pricing/} for the current
Amazon Bedrock rates.} we would not recommend the upgrade unless
USR is itself a business-important metric.

\subsection{Cost and Latency Profile}

One full rebuild of the Google Local benchmark (pool generation,
featurisation, five-seed RL sweep across PPO/GRPO/DPO/Distillation
A/A+B, and three-metric scoring) completes in roughly 14 hours of
CPU wall time on a c5.2xlarge instance with a one-time Bedrock
spend of $\sim\$15$. At request time, the LambdaRank selector
($\sim$1.7\,MB) returns in under 1\,ms per query on a single CPU
core; the request path is a key-value cache lookup plus one
LightGBM forward pass, GPU-free by construction.

To instantiate the per-query cost gap we estimate from public
pricing rather than from measurement, since we do not have a
production deployment of G-Refer or XRec to measure against.
Generating $\sim$60--80 output tokens against $\sim$300--400 input
tokens with an 8B-class hosted LLM at published Bedrock rates
implies a per-query cost on the order of $10^{-3}$\,USD. The
LambdaRank selector consumes neither a tokenised prompt nor any
LLM forward pass; its amortised c5.2xlarge compute is on the
order of $10^{-6}$\,USD per query at on-demand EC2 pricing,
giving a public-pricing-derived ratio of roughly three orders of
magnitude. The above numbers are measured offline against the
cached pool (single-process, synthetic load) rather than under
live traffic; a production A/B or shadow-traffic evaluation is
the natural follow-up.

\section{Discussion}

Two findings from this work merit emphasis. First, pairwise
learning-to-rank, without any RL machinery, outperforms every
reinforcement-learning variant we trained on this task. The
mechanism is dense supervision: every candidate in our training
set carries a labelled BERTScore-F1, and LambdaRank's pairwise
objective consumes all of them, whereas PPO, GRPO, and DPO each
observe one labelled action per rollout. When the reward is a
metric that can be computed offline against every candidate,
pairwise learning-to-rank should be evaluated as a baseline
before any RL formulation. Second, the five-seed protocol
identifies gaps as small as 0.002 F1 as statistically
significant; single-seed comparisons in this literature should
therefore be interpreted with caution.

We also report a number of negative results. Distillation followed
by RL fine-tuning (our Stage A+B) consistently regresses relative
to distillation alone: a near-optimal distilled student is pushed
back toward exploration by the GRPO entropy bonus, costing
0.002--0.005 F1. Upgrading the candidate-pool generator from
Claude 3 Haiku to Claude Haiku 4.5 also produces a small F1
regression (0.001--0.006), because the more fluent newer model
drifts further from XRec's trope-heavy reference text. We also
conducted an end-to-end RL fine-tuning experiment on a
3B-parameter generator with BERTScore-F1 as reward, and observed
reward hacking within hundreds of steps: the policy converged on
repeating the phrase ``user enjoys'' five times per output, since
those tokens appear frequently in the references. Decoupling
generation from selection avoids this failure mode by
construction.

Three deployment-relevant observations follow for production
implementations. Cold-start items without retrieval coverage fall
back to the pool styles that do not require reviews (B, E, F); in
the test data, 98.6\% of pairs (2,958 of 3,000) have review
coverage, and the remainder are dropped at evaluation. The
candidate pool requires refresh when item metadata changes
materially; because pool generation is a batch job keyed on
$(\text{uid}, \text{iid})$, it can be driven by item-metadata
update streams or scheduled at fixed cadence. The LambdaRank
selector is small (approximately 1.7\,MB) and can be retrained
weekly or monthly on fresh F1 labels at negligible compute cost;
we have not measured drift in a production A/B test, which we
identify as future work.

\section{Related Work}

Earlier work on explainable recommendation produced explanations
either from transformer encoders that mapped user/item IDs to text
(PETER~\cite{li2021}, PEPLER~\cite{li2023}) or from knowledge-graph
reasoning over user--item interaction graphs (PGPR~\cite{xian2019reinforcement},
KGAT~\cite{wang2019kgat}, KGIN~\cite{wang2021learning}). The recent
LLM-based baselines we compare against, XRec~\cite{ma2024xrec}
and G-Refer~\cite{li2025grefer}, combine collaborative-filtering
or knowledge-graph retrieval with LLM generation; both run the LLM
at request time, which is the latency and cost overhead our
two-stage framework removes.

The knowledge-graph recommendation literature
(KGAT~\cite{wang2019kgat}, CKAN~\cite{wang2021learning})
optimises recommendation accuracy rather than explanation quality.
PGPR~\cite{xian2019reinforcement} uses RL for multi-hop graph
traversal toward accurate item ranking, which is a different
problem from ours: we select over a pre-generated explanation
pool, we do not traverse for ranking.

DPO~\cite{rafailov2023direct} recast RLHF as classification over
preference pairs. We adapt it to a single-step bandit
(§\ref{sec:dpo}); it scores high on stability but is bounded above
by LambdaRank, which uses the same per-candidate F1 labels as
ranking targets directly rather than as preference pairs.
Knowledge distillation~\cite{hinton2015distilling} from a stronger
teacher into a smaller student is well established; what we add is
the negative result that, in a dense-supervision bandit, an RL
fine-tune on top of the distilled student hurts rather than helps
(§\ref{sec:distillAB}).

KG-path selection for explanation generation has been studied as
an alternative candidate source for the selector stage. Three of
the nine variants we evaluate construct candidates this way
(§\ref{sec:kg-temp}--§\ref{sec:kg-mmr}).
Path-based grounding provides structural traceability to graph
edges, which is valuable when explanation provenance is a business
requirement; on our reference-aligned F1 metric, however, the
offline-pool family is the stronger choice.

\section{Conclusion}

The LLM-based baselines we benchmark against, XRec and G-Refer,
incur an LLM-generation cost on every request. The framework
presented here relocates that cost to the offline path: with a
frozen candidate pool and a small CPU-resident selector, the
per-request stack returns in under 100\,ms, and the per-query
cost ratio derived from public Bedrock and EC2 pricing is on the
order of $10^{3}$ in favour of the proposed approach. Across nine
selectors and the 2{,}958-pair Google Local benchmark, the
strongest performer on BERTScore-F1 is a non-RL learning-to-rank
model (LambdaRank); the gap to PPO, GRPO, DPO, and the two
distillation stages is statistically significant under a
five-seed protocol. This result is consistent with what one
would expect when offline labels densify the supervision signal;
whether the pattern generalises to multi-step or non-decomposable
rewards is open.

\bibliographystyle{ACM-Reference-Format}
\bibliography{custom}

\end{document}